\documentclass{article}

\usepackage[T1]{fontenc}
\usepackage{microtype}
\usepackage{graphicx}
\usepackage{subfigure}
\usepackage{booktabs} 

\usepackage{hyperref}

\usepackage[accepted]{style/icml2025}

\usepackage{amsmath}
\usepackage{amssymb}
\usepackage{mathtools}
\usepackage{amsthm}

\usepackage[capitalize,noabbrev]{cleveref}
\theoremstyle{plain}

\theoremstyle{definition}

\theoremstyle{remark}

\usepackage[textsize=tiny]{todonotes}

\icmltitlerunning{Test short title ICML 2025}

\usepackage[inkscapelatex=false]{svg}
\date{}
\title{}
\hypersetup{
 pdfauthor={Alan Munoz},
 pdftitle={},
 pdfkeywords={},
 pdfsubject={},
 pdfcreator={Emacs 30.1 (Org mode 9.7.11)}, 
 pdflang={English}}
\usepackage{natbib}
\begin{document}

\twocolumn[
\icmltitle{cp\_measure: API-first feature extraction for image-based profiling workflows}



\icmlsetsymbol{equal}{*}

\begin{icmlauthorlist}
\icmlauthor{Al\'an F. Mu\~{n}oz}{broad}
\icmlauthor{Tim Treis}{hh,broad}
\icmlauthor{Alexandr A. Kalinin}{broad}
\icmlauthor{Shatavisha Dasgupta}{broad}
\icmlauthor{Fabian Theis}{hh}
\icmlauthor{Anne E. Carpenter}{broad}
\icmlauthor{Shantanu Singh}{broad}
\end{icmlauthorlist}

\icmlaffiliation{broad}{Broad Institute of MIT and Harvard, United States}
\icmlaffiliation{hh}{Institute of Computational Biology, Helmholtz Zentrum München, Germany}

\icmlcorrespondingauthor{Al\'an F. Mu\~{n}oz}{amunozgo@broadinstitute.org}
\icmlcorrespondingauthor{Shantanu Singh}{shantanu@broadinstitute.org}

\icmlkeywords{Machine Learning, ICML}

\vskip 0.3in
]



\printAffiliationsAndNotice{}  

\begin{abstract}
Biological image analysis has traditionally focused on measuring specific visual properties of interest for cells or other entities. 
A complementary paradigm gaining increasing traction is image-based profiling -- quantifying many distinct visual features to form comprehensive profiles which may reveal hidden patterns in cellular states, drug responses, and disease mechanisms.
While current tools like CellProfiler can generate these feature sets, they pose significant barriers to automated and reproducible analyses, hindering machine learning workflows. Here we introduce cp\_measure, a Python library that extracts CellProfiler's core measurement capabilities into a modular, API-first tool designed for programmatic feature extraction. We demonstrate that cp\_measure features retain high fidelity with CellProfiler features while enabling seamless integration with the scientific Python ecosystem. Through applications to 3D astrocyte imaging and spatial transcriptomics, we showcase how cp\_measure enables reproducible, automated image-based profiling pipelines that scale effectively for machine learning applications in computational biology.
\end{abstract}

\section{Introduction}
\label{sec:org4c9ba67}
High throughput screening of biological phenomena via complex modalities, such as RNA sequencing, is prohibitively expensive, therefore microscopy is an efficient first step. Through microscopy, modern biologists use a wide array of fluorescence dyes or proteins to observe the location and distribution of cells, organelles, and other components. Nowadays, quantification of biological images is standard, with software often identifying regions of interest (such as cells) and extracting features that represent these regions, such as intensity.

Image-based profiling--also termed morphological profiling--is a technique of measuring an extensive suite of morphological features for a population of biological objects, such as cells. 
These features are fed into statistical or machine learning methods to identify biologically meaningful patterns. One of its biggest applications of morphological profiling is drug discovery, where scientists leverage microscopy's low acquisition cost and high throughput to accomplish many goals, such as grouping genes by function, identifying chemical compounds that target a protein, and predicting toxicity of drug candidates \citep{chandrasekaranImagebasedProfilingDrug2021}. 

\begin{figure}[htbp]
\centering
\includesvg[width=.99\linewidth]{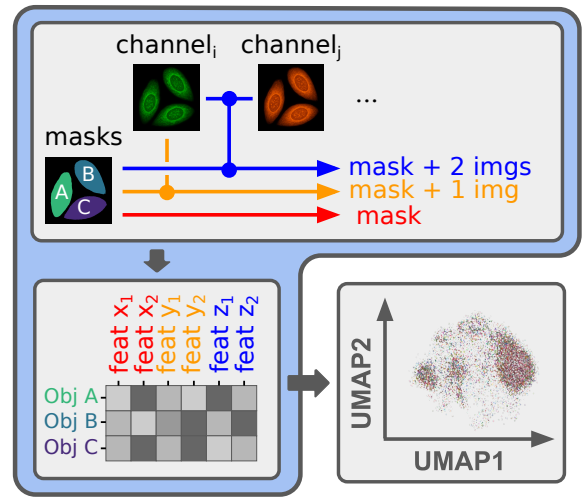}
\caption{\label{fig:overview}cp\_measure generates features from images by using information in every region of interest ("object"). It can featurize the pairwise combination of all the available channels (colours) and objects. The resultant matrices represent the entire experiment and can be studied using statistical, machine, and deep learning methods.}
\end{figure}

\subsection{The current state of bioimage analysis}
\label{sec:org8f5b33d}
The most widely used software for processing high-throughput biological images is CellProfiler \citep{stirlingCellProfiler4Improvements2021}, providing experimental biologists with limited programming expertise an accessible yet powerful toolset. By contrast, it may not suit the needs of computational biologists building high throughput pipelines comprised of multiple tools. CellProfiler is ideal for creating and iteratively adapting manually-defined workflows; for low-customization high-throughput analyses such as extracting features for image-based profiles or spatial transcriptomics data, however, certain aspects become a hindrance. For example, complementing CellProfiler with other image analysis tools is a struggle, as building plug-ins to add functionality is a time and effort-consuming challenge that requires an understanding of its interfaces. Lastly, CellProfiler depends on many Python and Java packages, increasing the likelihood of dependency conflicts. Containers mitigate this problem, but not without increased complexity and caveats.

Alternative tools can extract features for image-based profiles, such as scikit-image \citep{waltScikitimageImageProcessing2014}, ScaleFex, or SpaCR \citep{comoletHighlyEfficientScalable2024,einarolafssonSpaCr2025}, however, these approaches present trade-offs for different use cases. Their independent implementation means they generate different feature sets than CellProfiler, which can be advantageous for novel
 applications but complicates direct comparison with established datasets like those in the Cell Painting Gallery \cite{weisbartCellPaintingGallery2024}. 
Additionally, while all tools necessarily impose input requirements, some enforce rigid organizational constraints--demanding specific directory structures and naming conventions--that require extensive data reorganization before analysis, beyond the standard data type and format specifications.
Some alternatives also target specific deployment scenarios, with ScaleFex designed primarily for cloud environments, limiting utility for researchers preferring local computation.

\subsection{Engineered features in the era of deep learning}
\label{sec:org9dc3dfa}

While computer vision has largely transitioned to deep learning-based feature extraction, biological imaging presents unique challenges that maintain the relevance of engineered features. These images differ substantially from natural images in their acquisition methods, visual properties, and analytical objectives, often limiting the effectiveness of standard pre-trained models, even those trained on biological data.

The application of deep learning to feature extraction for biological image analysis shows mixed results.
When appropriately trained on domain-specific data, deep learning networks can outperform engineered features \cite{lafargeCapturingSingleCellPhenotypic2019,moshkovLearningRepresentationsImagebased2022,chowPredictingDrugPolypharmacology2022,wolfSCANPYLargescaleSinglecell2018}, but not universally \cite{tangMorphologicalProfilingDrug2024,kimSelfsupervisionAdvancesMorphological2023}. 
Effective deep learning feature extractors for biological applications typically require substantial domain-specific datasets and careful model architecture choices, rather than simple transfer from general computer vision models.

Several practical considerations favour engineered features in biological imaging. 
First, interpretability remains crucial for biological discovery--researchers need to understand which morphological characteristics drive observed phenotypes. 
While deep learning features can be powerful predictors, their interpretation requires sophisticated techniques that may not provide the mechanistic insights biologists seek \citep{liChallengesOpportunitiesBioimage2023}. 
Engineered features have clear mathematical definitions that directly translate to biological concepts, such as protein co-localisation or nuclear morphology changes \cite{garcia-fossaInterpretingImagebasedProfiles2023}.

Second, many biological imaging laboratories lack the GPU resources, large annotated datasets, or machine learning expertise required for custom model development. While foundational models are emerging, visual foundational models can be challenged by the diversity of imaging modalities and experimental conditions \cite{azadFoundationalModelsMedical2023}.

Third, engineered features offer consistency and reproducibility advantages valuable for comparative studies and multi-site collaborations, producing identical results across different computing environments.

Rather than viewing these approaches as competing alternatives, the field increasingly recognizes their complementary roles. 
This motivates the need for robust, accessible tools that efficiently extract interpretable features while integrating seamlessly with modern computational workflows.
With all this in mind, we developed cp\_measure.

\section{Extracting interpretable features}
\label{sec:org61842b5}
The library cp\_measure branches off the CellProfiler codebase, adapted to include calculations for all features of an image-based profiling pipeline, while removing the user interface and anything else non-essential to the task. Measurements are defined as a collection of related features, following CellProfiler's nomenclature, and are categorized based on three kinds of input: One object (producing features such as area and eccentricity), one object + one imaging channel (these can be used to extract such measurements as intensity, texture), one object + multiple channels (e.g., for Manders correlations between intensity values across channels). A visual representation of this can be found in Figure~\ref{fig:overview}.

By isolating the feature calculations from the graphical interface and orchestration components, we aimed to make them accessible to the data science community while improving reusability, testability and extensibility. This reduces the amount of time and human effort required to integrate featurization into existing and new pipelines, while still retaining the standard set of features available for multiple public datasets.

\subsection{Reproducing CellProfiler measurements}
\label{sec:org09b0cd2}
\begin{figure}[htbp]
\centering
\includesvg[width=.9\linewidth]{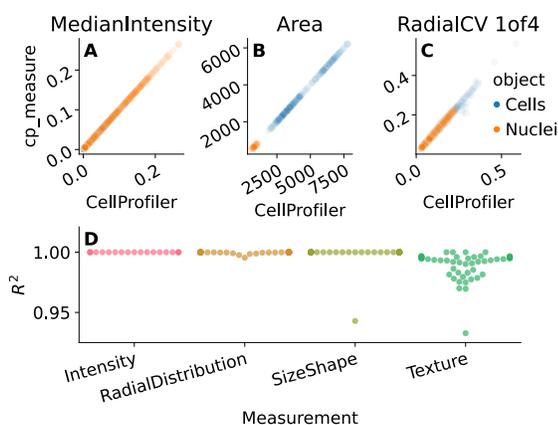}
\caption{\label{fig:cp_vs_cpmeasure}cp\_measure features match their CellProfiler analogues. \textbf{Panels A-C.} Representative examples comparing CellProfiler feature values (x-axis) to cp\_measure's (y-axis), generated using matching pairs of masks and images. \textbf{Panel D.} \(R^2\) value of a linear fit for each individual feature, comparing cp\_measure to CellProfiler.}
\end{figure}

We first tested whether cp\_measure features match the original CellProfiler features, for a representative set of 300 images. These correspond to 150 perturbations from the JUMP Cell Painting dataset \citep{chandrasekaranJUMPCellPainting2023}, selecting a representative subset of the most significant phenotypes for each feature. We segmented images to delineate the cells and nuclei using CellProfiler, providing object masks (regions of interest) to CellProfiler and to cp\_measure for feature extraction. Next, we applied cp\_measure on these masks and the original images. 
To validate cp\_measure, we compared its output directly with CellProfiler's on identical images and masks. 
For each feature, we computed the $R^2$ value between the two tools' measurements. 
Over 95\% of features showed $R^2$ > 0.9 (Figure \ref{fig:cp_vs_cpmeasure}), indicating near-perfect agreement. 
The few outliers with lower correlations reveal edge cases handled differently by each tool; these will be addressed through comprehensive unit testing. These results indicate that cp\_measure is not likely to produce massively different profiles than those calculated by CellProfiler.

\subsection{Results}
\label{sec:orge5b5c6b}
\subsubsection{Astrocyte 3D nuclei data}
\label{sec:org447090b}

\begin{figure}[htbp]
\centering
\includesvg[width=.9\linewidth]{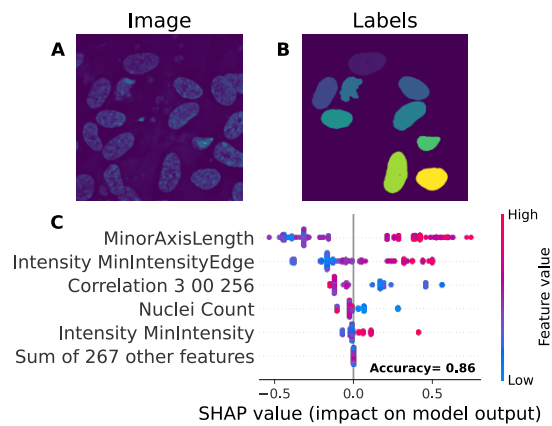}
\caption{\label{fig:astrocytes}\textbf{Panels A and B.} Example pair of astrocyte nuclei image and masks. The 3D images were projected over the z-axis, taking the maximum value across the z-stack. \textbf{Panel C.} SHAP values of the most important features for classifying the day of differentiation (out of three days). Each point represents a single 3D image containing multiple nuclei, the magnitude of SHAP values indicate how important those features are for the classifier. Features are ranked by importance, with individual SHAP values shown for the top features and the summed impact shown for the remaining 267 features shown as a single aggregated value. Shape features, such as the minor axis length, refer to the nucleus, while intensity features refer to the DNA staining. The test data accuracy is shown in bold.}
\end{figure}

We demonstrate cp\_measure's utility through a biologically relevant task: tracking astrocyte maturation over time. 
From 433 3D images containing 831 astrocyte nuclei \citep{kalininValproicAcidinducedChanges2021}, cp\_measure extracted morphological features  that we preprocessed following standard procedures \citep{caicedoDataanalysisStrategiesImagebased2017}.
We then trained a Gradient Boosting classifier to predict the day of cell differentiation with 87\% accuracy. 
SHAP analysis \citep{sundararajanManyShapleyValues2020} (\ref{fig:astrocytes}) revealed that nuclear minor axis length was the most predictive feature: nuclei became progressively wider during maturation, a phenotype consistent with known astrocyte biology.

\subsubsection{Spatial transcriptomics}
\label{sec:org5711d86}
To demonstrate the broad applicability of cp\_measure, we developed a morphology featurizer for the popular spatial omics analysis toolbox Squidpy~\citep{pallaSquidpyScalableFramework2022}.
Many of the recently developed spatial omics analysis technologies generate, in addition to their target omics layer, imaging data, usually H\&E (hematoxylin and eosin) stained or fluorescently labelled. 

Spatial transcriptomics poses unique computational challenges: images routinely contain 500,000+ cells, far exceeding CellProfiler's memory constraints and rigid data structure requirements. 
cp\_measure's modular design enabled us to build a custom workflow integrated with Squidpy~\citep{pallaSquidpyScalableFramework2022} that leverages the SpatialData format \citep{Marconato2024-SpatialData} to process cells in streaming parallel batches. 
This architecture handles massive spatial datasets efficiently while bringing morphological profiling directly into the scverse ecosystem \citep{Virshup2023-scverse}, making these features immediately accessible to the single-cell community.

To illustrate the value of this workflow, we used it to identify morphologically similar regions within two human breast cancer sections profiled with the spatial transcriptomics technology Xenium \citep{10x-Genomics2023-el}. We generated a pattern of tessellating hexagonal masks across both images, of which 18469 were found to overlay the tissue. These masks were then featurized with a cp\_measure-based workflow: We first preprocessed the features following standard workflows
\citep{serranoReproducibleImagebasedProfiling2025}, then removed batch effects between the samples using Harmony \citep{Korsunsky2019-Harmony}, and finally obtained the Leiden clusters via Scanpy \citep{wolfSCANPYLargescaleSinglecell2018, Traag2019-fr}. 

\begin{figure}[htbp]
\centering
\includegraphics[width=.99\linewidth]{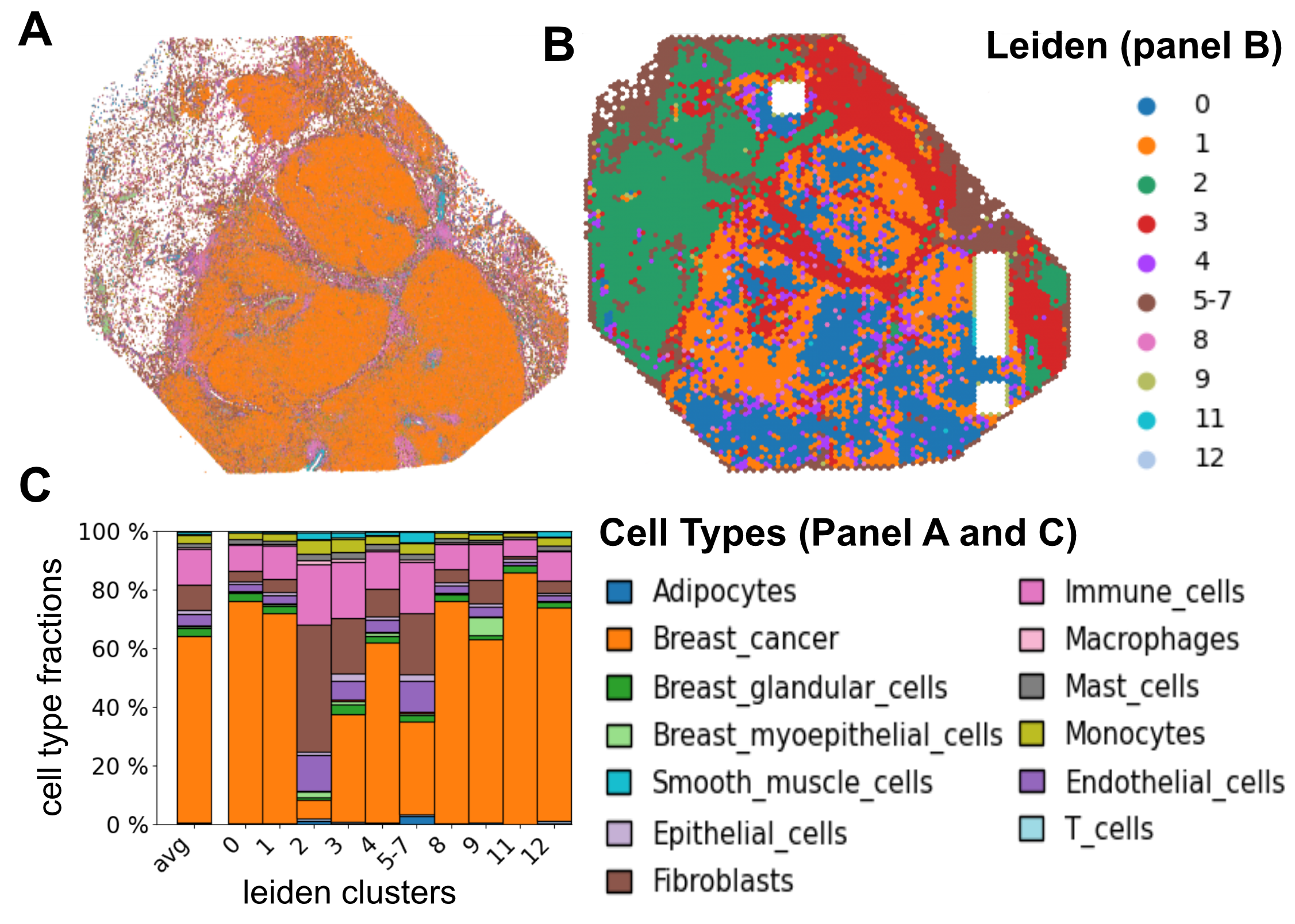}
\caption{\label{fig:spatial_omics}{}Morphology reveals the composition of tumour subregions. \textbf{Panel A.} Spatial map of cell types inferred from transcriptomics. \textbf{Panel B.} Morphological subregions obtained by Leiden clustering the image-derived feature embeddings. \textbf{Panel C.} Bar chart showing, for each morphological cluster, the fraction of cells assigned to each cell type. Clusters 5-7 were merged due to similarity, cluster~10 removed due to low cell count (see Figure \ref{fig:spatial_c10}B).}
\end{figure}

For each cluster, we extracted the cells within its hexagons and calculated the respective cell type proportions. A visual comparison of panels A and B in Figure~\ref{fig:spatial_omics} shows that morphology-only clustering largely recapitulates the underlying tissue architecture. Panel C reveals that some clusters have distinct cell-type compositions: For example, cluster~2 is dominated by fibroblasts, whereas adipocytes appear almost exclusively in clusters 5-7. Furthermore, comparing clusters~0 and 1 suggest intra-tumour differences, which could be correlated to immune-hotness (see Figure \ref{fig:spatial_immunehot}).

These results demonstrate how cp\_measure can can be used outside traditional microscopy. When applied on extremely large images it produces biologically meaningful morphological clusters that validate and complement transcriptomic phenotypes. By embedding this capability within the scverse ecosystem via Squidpy, we enable researchers to seamlessly add morphology to their spatial omics workflows, unlocking new insights into tissue organization.

\section{Discussion}
\label{sec:orgf37b369}
While point-and-click interfaces open the world of image analysis to many researchers, they are not as effective for computational workflows with no human-in-the-loop. In this work we introduced our new library cp\_measure, which calculates a set of widely used engineered features relevant to whole images and to regions of interest (object masks). It enables simpler automated profiling of microscopy data in short scripts and complex pipelines. The modularity it provides facilitates pipelines with better scaling capabilities for high-content microscopy, with or without cloud infrastructure.

The biologically interpretable features provided by cp\_measure complement deep learning features and offer a better mechanistic understanding of the underlying biology. A potential workflow would be to use deep learning features to cluster and then use cp\_measure features from these clusters. As a whole, when used in tandem with generalist image-processing tools, such as Cellpose for segmentation \cite{stringerCellposeGeneralistAlgorithm2021}, machine and deep learning workflows can be streamlined. 

While developed for biology, engineered morphological features have proven valuable far beyond their original domain--from analysing microplastics in environmental samples \citep{ideharaExploringNileRed2025} to quality control in manufacturing \citep{ilhanEffectsProcessParameters2021} and agriculture automation \citep{xuDesignNonDestructiveSeed2024}. 
cp\_measure's lightweight, API-first design makes these powerful features accessible to any field working with image data, removing the barriers of biological-specific interfaces.

\section{Future work}
\label{sec:org5cdbb12}
We propose to make cp\_measure an imported dependency for CellProfiler, offering several benefits. It would ensure that the results from pipelines built with either tool will always be comparable, while also providing the opportunity to formalize the inputs and outputs of all measurements. It would also mean that any changes made in cp\_measure propagate to CellProfiler, benefiting the community for which it was originally developed.

We also plan to develop a comprehensive test suite to guarantee mathematical correctness, which currently CellProfiler itself is lacking. A comprehensive test suite would enable more confident and expedient optimization for the most compute-consuming features, such as granularity, by providing rapid iteration capabilities. Once tests are in place, we could add support for just-in-time compiling and GPUs. We envision that cp\_measure could also be the place to develop and distribute new measurements from the community.

\section{Acknowledgements}
\label{sec:acknowledgements}

This work was supported by GSK and the National Institutes of Health, grant (R35 GM122547 to AEC). The authors would like to thank Nodar Gogoberidze, Beth Cimini, Minh Doan and Eliot McKinley for their valuable feedback and discussions during the course of this project.

\subsection{Methods}
\label{sec:orgb3e9382}
\subsubsection{Data and software}
\label{sec:orgbda0ae2}
The codebase for cp\_measure is available on \url{https://github.com/afermg/cp_measure}. All code to reproduce the analyses and figures, alongside links to the original data, is available on the GitHub repository \url{https://github.com/afermg/2025_cpmeasure}.


\bibliographystyle{icml2025}
\bibliography{bibliography}

\onecolumn
\section{Appendix}

\appendix
\setcounter{figure}{0}
\renewcommand{\thefigure}{A\arabic{figure}}

\setcounter{section}{5}
\renewcommand{\thesection}{\arabic{section}}
\renewcommand{\thesubsection}{\thesection.\arabic{subsection}}
\renewcommand{\thesubsubsection}{\thesubsection.\arabic{subsubsection}}

\subsubsection{Details on the spatial use case}
\label{sec:spatial_extra}

\begin{figure}[htbp]
\centering
\includegraphics[width=.8\linewidth]{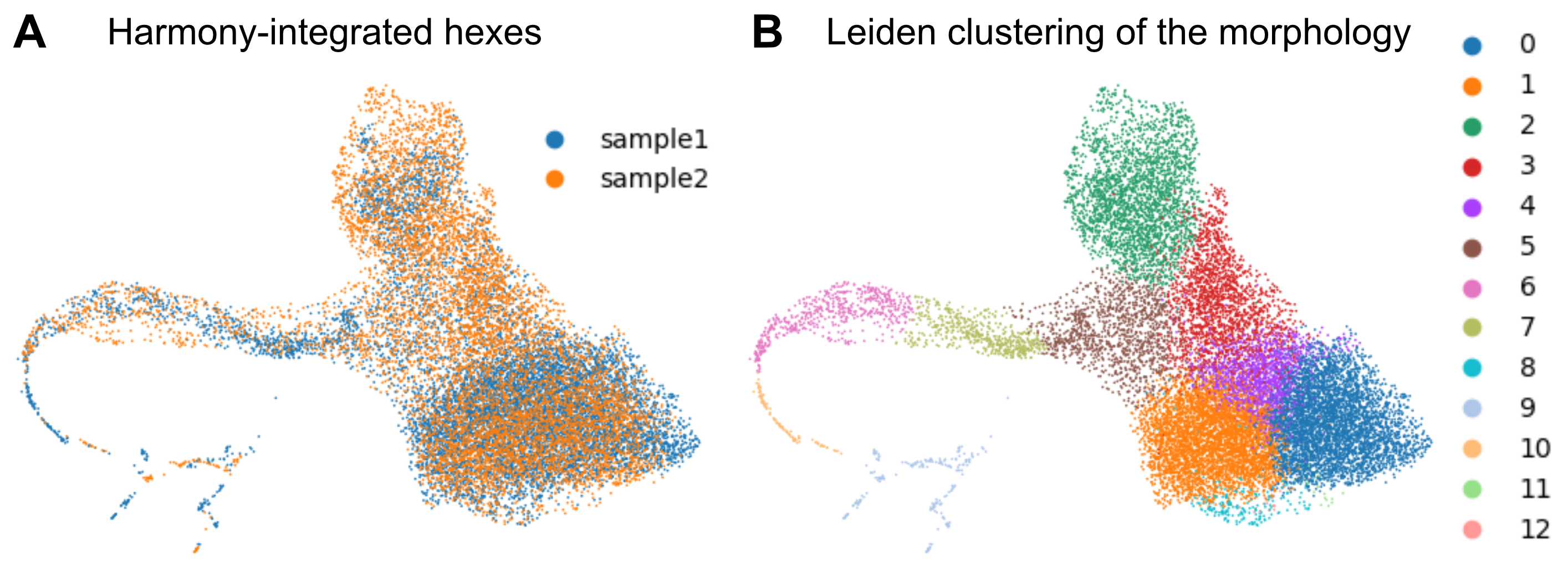}
\caption{\label{fig:spatial_umap}{}UMAP projection of the processed morphology features. \textbf{Panel A.} Coloured by sample of origin, we see good mixing after the Harmony integration. \textbf{Panel B.} Coloured by Leiden clustering.}
\end{figure}

\begin{figure}[htbp]
\centering
\includegraphics[width=.65\linewidth]{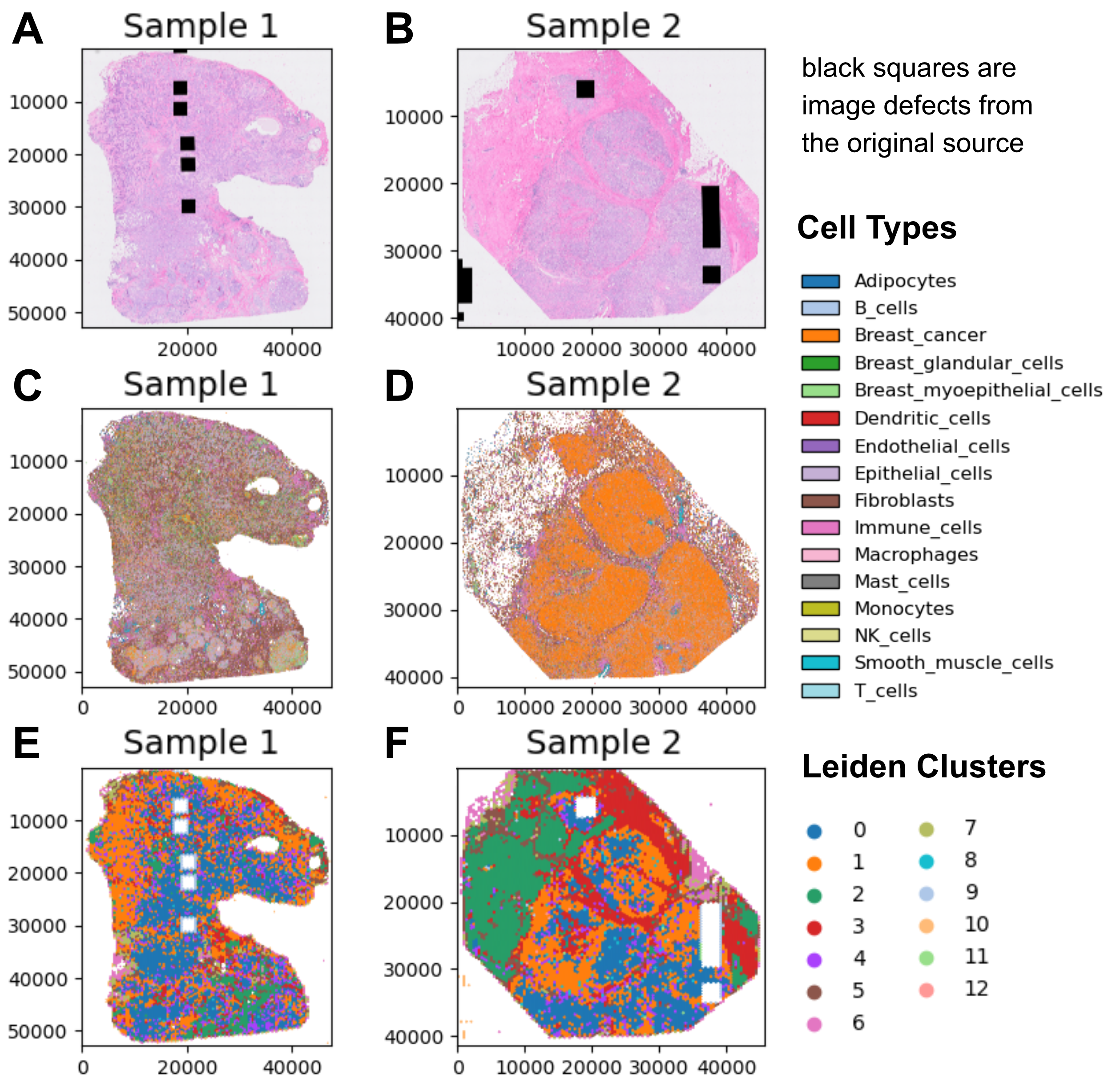}
\caption{\label{fig:spatial_hne}{}Overview about the spatial distribution of cell types and morphology clusters. \textbf{Panel A and B.} H\&E images of both samples. \textbf{Panel C and D.} Cell segmentation masks colored by cell type. \textbf{Panel E and F.} Hexagons colored by their morphology Leiden cluster.}
\end{figure}

The following figures show the processing steps for the spatial data used to generate Figure~\ref{fig:spatial_omics} in more detail. Raw data from two Xenium Spatial Transcriptomics available from the 10x Genomics Datasets homepage (name: "FFPE Human Breast with Custom Add-on Panel", release date: 2023-01-22) were downloaded and converted to the `SpatialData` format. The data comprises two samples, one infiltrating ductal carcinoma ("sample 1") and one invasive lobular carcinoma ("sample 2"). H\&E images from the same dataset were also downloaded, aligned to the respective reference images, and saved to the data object. These can be seen in Figure~\ref{fig:spatial_hne}A and Figure~\ref{fig:spatial_hne}B. A pattern of tessellating hexagons was generated for each sample, spanning the entire tissue section. These hexagons were subselected to only those overlaying the specimens as visible in the H\&E image, resulting in 10321 hexagons for sample~1 and 8148 for sample~2 respectively. These hexagonal masks were used to automatically extract the underlying H\&E crops which were then featurized using `Squidpy`, resulting in 945 features each. These features were then pre-processed using `Pycytominer`, resulting in 252 features. We then removed batch effect using Harmony. UMAP projections of these features are shown in Figure~\ref{fig:spatial_umap}.

\begin{figure}[htbp]
\centering
\includegraphics[width=.7\linewidth]{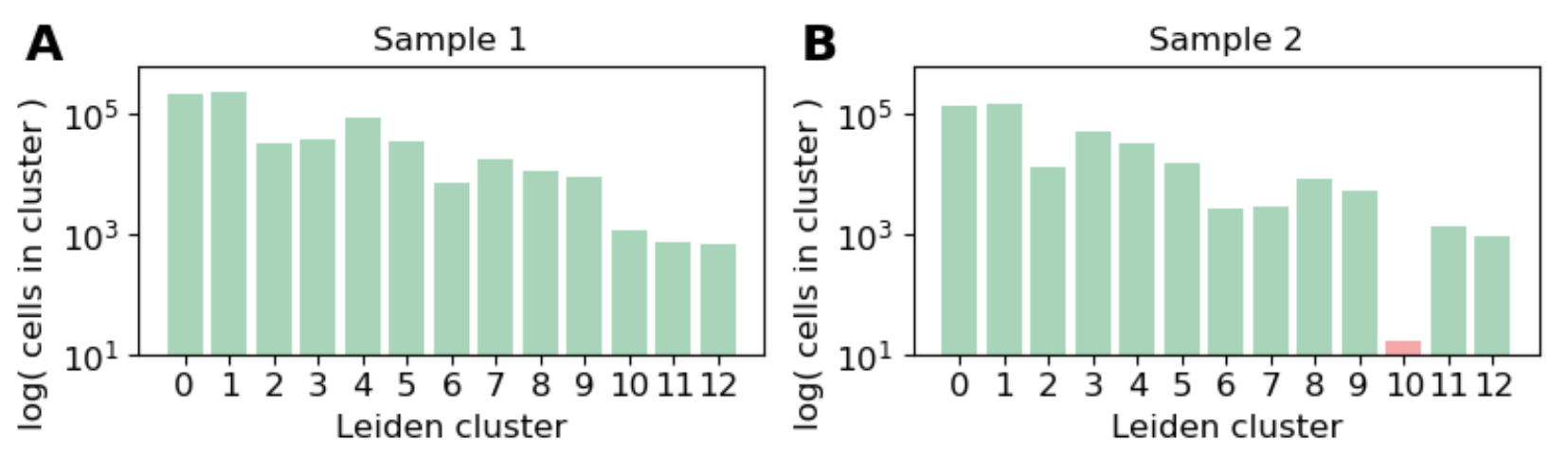}
\caption{\label{fig:spatial_c10}{}Morphology reveals the composition of tumour subregions. \textbf{Panel A.} Spatial map of cell types inferred from transcriptomics. \textbf{Panel B.} Morphological subregions obtained by applying Leiden clustering to image-derived feature embeddings. \textbf{Panel C.} Bar chart showing, for each morphological cluster, the fraction of cells assigned to each cell type (colour legend as shown). The top left quarter of Panel A shows that the morphological approach worked to predict regions while the expression annotations did not.}
\end{figure}

Due to the low cell count in Leiden cluster 10 for sample 2, as shown in Figure~\ref{fig:spatial_c10}B the cluster was excluded for some analysis steps.

\newpage

Then, the gene expression data from the Xenium experiment was used to annotate cell types for both samples. For every Leiden cluster, as defined by the Leiden clustering of the joint morphology features from both samples, the cells covered by the respective hexagons were extracted. Figure~\ref{fig:spatial_hne} shows the raw H\&E images, the distribution of cell types across these, and the hexagons coloured by their respective Leiden cluster. Of note is that the H\&E images both seemed to be partially corrupted, resulting in blacked-out sections. While the data contains cell segmentation masks for these sections, we filtered out hexagons overlaying these corrupted regions for the morphology featurization.

\begin{figure}[htbp!]
\centering
\includegraphics[width=.95\linewidth]{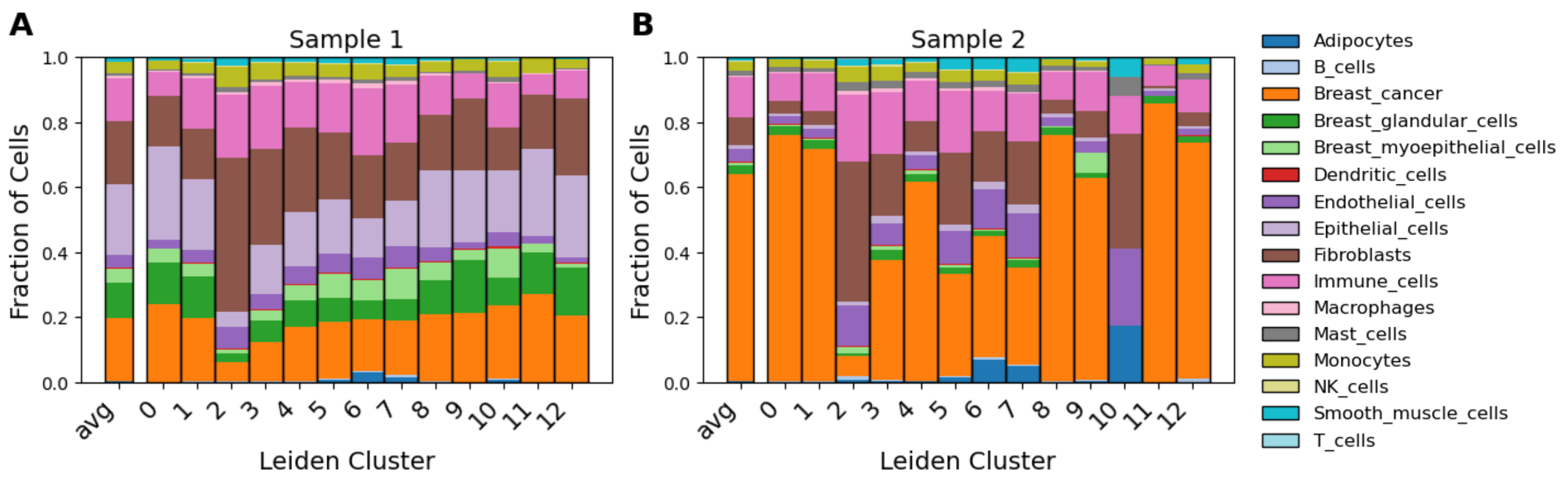}
\caption{\label{fig:spatial_composition}{}Barplots showing the per-cluster cell type composition for both samples. Additionally, the average per sample is shown.}
\end{figure}

For every morphology Leiden cluster, we counted all cells whose centroids fell into the respective hexagons and calculated each cell type's share of the total cells per cluster. This data is shown in Figure~\ref{fig:spatial_composition}.

\newpage

\begin{figure}[htbp!]
\centering
\includegraphics[width=.85\linewidth]{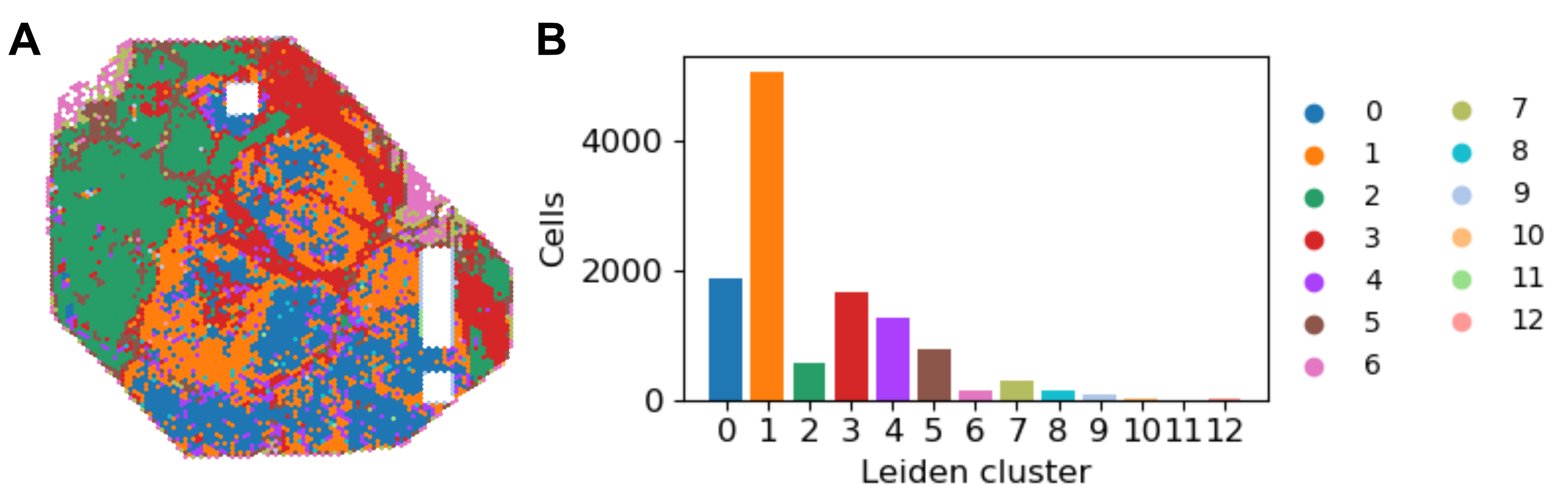}
\caption{\label{fig:spatial_immunehot}{}Potential explanation of the differences between cluster~0 and cluster~1. \textbf{Panel A} Hexagons in sample~2 coloured by Leiden cluster. \textbf{Panel B} Cumulative counts of B~cells, dendritic~cells, NK~cells, and T~cells.}
\end{figure}

Based on \citep{Wu2024-zp} we define an immune-hot signature comprising B~cells, dendritic~cells, NK~cells, and T~cells. For every Leiden cluster, we aggregated the respective cell type counts of this signature for both samples. These are shown in Figure~\ref{fig:spatial_immunehot}B and provide initial evidence for the difference in clusters 0 and 1, despite both being visually mixed as can be seen in Figure~\ref{fig:spatial_immunehot}A.

\end{document}